\documentclass[10pt,twocolumn,letterpaper]{article}

\usepackage{cvpr}
\usepackage{times}
\usepackage{epsfig}
\usepackage{graphicx}
\usepackage{amsmath}
\usepackage{amssymb}

\usepackage{paralist}
\usepackage{comment}
\usepackage{xcolor}
\usepackage{cite}

\newcommand{\vgp}{\textsc{VGPhraseCut}\xspace}

\definecolor{citecolor}{HTML}{0071bc}
\usepackage[pagebackref=false,breaklinks=true,letterpaper=true,colorlinks,citecolor=citecolor,bookmarks=false]{hyperref}
\cvprfinalcopy 

\def\cvprPaperID{781} 

\ifcvprfinal\fi

\begin{document}


\title{PhraseCut: Language-based Image Segmentation in the Wild} 
\author{
Chenyun Wu$^1$ \quad
Zhe Lin$^2$ \quad
Scott Cohen$^2$ \quad
Trung Bui$^2$ \quad
Subhransu Maji$^1$\\
$^1 $University of Massachusetts Amherst \quad
$^2 $Adobe Research \\
{\tt\small \{chenyun,smaji\}@cs.umass.edu,~
 \{zlin,scohen,bui\}@adobe.com}
}
 
 
\maketitle
\thispagestyle{empty}

\begin{abstract}
We consider the problem of segmenting image regions given
a natural language phrase, and study it on a novel dataset of 77,262 images and 345,486 phrase-region pairs.
Our dataset is collected on top of the Visual Genome dataset and uses the
existing annotations to generate a challenging set
of referring phrases for which the corresponding regions are manually annotated.
Phrases in our dataset correspond to multiple regions and describe a large number of object and stuff categories as well as their attributes such as color, shape, parts, and relationships with other entities in the image.
Our experiments show that the scale and diversity of concepts in our
dataset poses significant challenges to the existing state-of-the-art.
We systematically handle the long-tail nature of these concepts and present a modular approach to combine category, attribute, and relationship cues that outperforms existing approaches. 

%

\end{abstract}

\vspace{-0.05in}
\section{Introduction}
\label{sec:intro}
Modeling the interplay of language and vision is important for tasks such as visual question answering, automatic image editing, human-robot interaction, and more broadly towards the goal of general Artificial Intelligence. 
Existing efforts on grounding language descriptions to images have achieved promising results on datasets such as \textit{Flickr30Entities}~\cite{plummer2017flickr30k} and \textit{Google Referring Expressions}~\cite{mao2016generation}.
These datasets, however, lack the scale and diversity of concepts that appear in real-world applications.

\begin{figure}
\centering
\includegraphics[width=\linewidth]{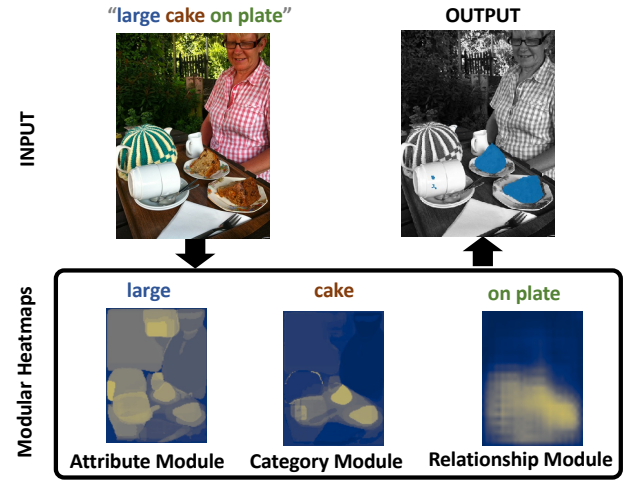}
\caption{\textbf{Our task and approach.} PhraseCut is the task of segmenting image regions given a natural language phrase. Each phrase is templated into words corresponding to \emph{categories}, \emph{attributes}, and \emph{relationships}. 
Our approach combines these cues in a modular manner to estimate the final output.
}
\vspace{-0.1in}
\label{fig:teaser}
\end{figure}

\begin{figure*}
\centering
\includegraphics[width=\linewidth]{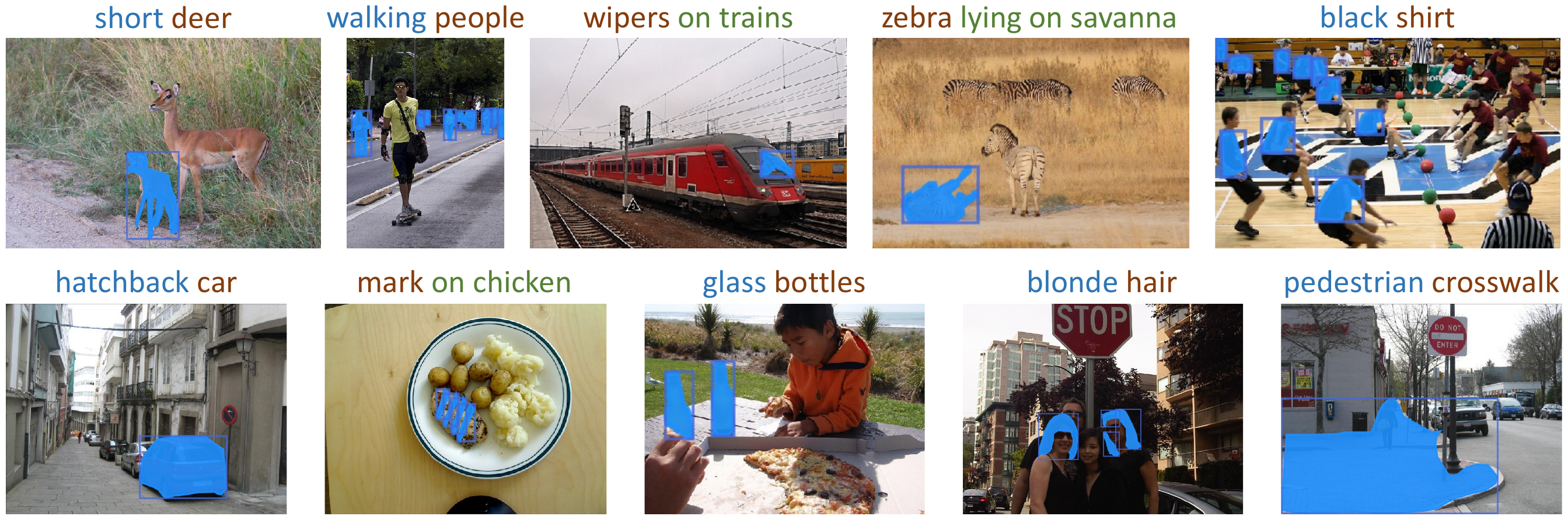}
\caption{\textbf{Example annotations from the \vgp dataset}. Colors (blue, red, green) of the input phrases correspond to words that indicate attributes, categories, and relationships respectively.}
\vspace{-0.1in}
\label{fig:eg}
\end{figure*}

To bridge this gap we present the \vgp dataset and an associated task of grounding natural language phrases to image regions called \emph{PhraseCut} (Figure~\ref{fig:teaser} and \ref{fig:eg}).
Our dataset leverages the annotations in the \textit{Visual Genome~(VG)} dataset~\cite{krishnavisualgenome} to generate a large set of referring phrases for each image.
For each phrase, we annotate the regions and instance-level bounding boxes that correspond to the phrase.
Our dataset contains 77,262 images and 345,486 phrase-region pairs, with some examples shown in Figure~\ref{fig:eg}.
\vgp contains a significantly longer tail of concepts and has a unified treatment of stuff and object categories, unlike prior datasets.
The phrases are structured into words that describe categories, attributes, and relationships, providing a systematic way of understanding the performance on individual cues as well as their combinations.


The \emph{PhraseCut} task is to segment regions of an image given a \emph{templated phrase}.
As seen in Figure~\ref{fig:teaser}, this requires connecting natural language concepts to image regions.
Our experiments shows that the task is challenging for state-of-the-art referring approaches such as \textit{MattNet}~\cite{yu2018mattnet} and \textit{RMI}~\cite{liu2017recurrent}.
We find that the overall performance is limited by the performance on rare categories and attributes.
To address these challenges we present (i) a modular approach for combining visual cues related to categories, attributes, and relationships, and (ii) a systematic approach to improving the performance on rare categories and attributes by leveraging predictions on more frequent ones. 
Our category and attribute modules are based on detection models, whose instance-level scores are projected back to the image and further processed using an attention-based model driven by the query phrase.
Finally, these are combined with relationship scores to estimate the segmentation mask (see Figure~\ref{fig:teaser}).
Objects and stuff categories are processed in a unified manner and the modular design, after the treatment of rare categories, outperforms existing end-to-end models trained on the same dataset. 

Using the dataset we present a systematic analysis of the performance of the models on different subsets of the data. The main conclusions are: (i) object and attribute detection remains poor on rare and small-sized categories, (ii) for the task of image grounding, rare concepts benefit from related but frequent ones (\eg, the concept ``policeman" could be replaced by ``man" if there were other distinguishing attributes such as the color of the shirt), and (iii) attributes and relationship models provide the most improvements on rare and small-sized categories.
The performance on this dataset is far from perfect and should encourage better models of object detection and semantic segmentation in the computer vision community. The dataset and code is available at: \url{https://people.cs.umass.edu/~chenyun/phrasecut}.


\vspace{-0.05in}
\section{Related Work}
\vspace{-0.05in}
\label{sec:related}
The language and vision community has put significant effort into relating words and images. 
Our dataset is closely related to datasets for the visual grounding of referring expressions. 
We also describe recent approaches for grounding natural language to image regions.

\begin{table*}
  \begin{center}
    \resizebox{\linewidth}{!}{
      \begin{tabular}{c|c|c|c|c|c|c}
	    \textbf{Dataset}
        & \textbf{ReferIt}~\cite{kazemzadeh2014referitgame} 		& \textbf{Google RefExp}~\cite{mao2016generation}
        & \textbf{RefCOCO}~\cite{yu2016modeling}  	        & \textbf{Flickr30K Entities}~\cite{plummer2017flickr30k}
        & \textbf{Visual Genome}~\cite{krishnavisualgenome}     & \textbf{\vgp}\\ 
        \hline
	\# images
        &19,894 			&26,711
        & 19,994	        & 31,783
        & 108,077 			& 77,262\\ 
        \# instances
	& 96,654 & 54,822
	& 50,000  & 275,775 
        & 1,366,673      & 345,486  \\ 
	\# categories
        & -				& 80
        & 80 				& 44,518
        &  80,138				&
        3103\\ 
	multi-instance  	&No &No &No &No & No & Yes \\ 
	segmentation  	&Yes&Yes& Yes&No& No & Yes \\ 
        
	referring phrase  &short phrases & long descriptions & short phrases & entities in captions  & region descriptions & templated phrases 

    \end{tabular}}
  \end{center}
  \vspace{-0.1in}
  \caption{\textbf{Comparison of visual grounding datasets.} The proposed \vgp dataset has a significantly higher number of categories than RefCOCO and Google RefExp, while also containing multiple instances.}
  \vspace{-0.1in}
  \label{tab:rel-dataset}
\end{table*}

\vspace{-0.15in}
\paragraph{Visual grounding datasets} 
Table~\ref{tab:rel-dataset} shows a comparison of various datasets
related to grounding referring expressions to images.
The \emph{ReferIt} dataset~\cite{kazemzadeh2014referitgame} was collected on images from ImageCLEF using a ReferItGame between two players.
Mao \etal~\cite{mao2016generation} used the same strategy to collect a significantly larger dataset called \emph{Google RefExp}, on images
from the MS COCO dataset~\cite{lin2014microsoft}.
The referring phrases describe objects and refer to boxes inside the image across
$80$ categories, but the descriptions are long and perhaps
redundant.
Yu \etal~\cite{yu2016modeling} instead collect referring expressions
using a pragmatic setting where there is limited interaction time
between the players to generate and infer the referring object. 
They collected two versions of the data:
\emph{RefCOCO} that allows location descriptions such as ``man on the
left", and \emph{RefCOCO+} which forbids location cues forcing a focus
on other visual clues.
One drawback is that \emph{Google RefExp}, \emph{RefCOCO} and \emph{RefCOCO+} are all
collected on \emph{MS-COCO} objects, limiting their referring targets to 80 object categories. 
Moreover, the target is always one single instance, and there is no
treatment of stuff categories.

Another related dataset is the \emph{Flickr30K Entities}~\cite{plummer2017flickr30k}.
Firstly entities are mined and grouped (co-reference resolution) from captions by linking phrases that describe the same entity and then the corresponding bounding-boxes are collected.
Sentence context is often needed to ground the entity phrases to image regions.
While there are a large number of categories (44,518), most of them have very few examples (average 6.2 examples per category) with a significant bias towards human-related categories (their top 7 categories are ``man",``woman", ``people", ``shirt", ``girl", ``boy", ``men").
The dataset also does not contain segmentation masks. nor phrases that describe multiple instances.
 

Our dataset is based on the \emph{Visual
  Genome (VG)} dataset~\cite{krishnavisualgenome}.
VG annotates each image as a ``scene graph'' linking
descriptions of individual objects, attributes, and their relationships to other objects in the image. 
The dataset is diverse, capturing various object and
stuff categories, as well as attribute and relationship types.
However, most descriptions do not distinguish one object from other objects in the scene, \ie, they are not referring expressions.
Also, VG object boxes are very noisy.
We propose a procedure to mine descriptions within the scene graph that uniquely identifies the objects, thereby generating phrases that are more suitable for the referring task.
Finally, we collect segmentation annotations of corresponding regions for these phrases.

\vspace{-0.15in}
\paragraph{Approaches for grounding language to images}
Techniques for localizing regions in an
image given a natural language phrase can be broadly categorized into two groups:
single-stage segmentation-based techniques and two-stage
detection-and-ranking based techniques.

Single-stage methods~\cite{hu2016segmentation, liu2017recurrent, li2018rrn, shi2018keyword, margffoy2018dynamic, ye2019cross, Chen_2019_ICCV, Yang_2019_ICCV} predict a segmentation mask given a
natural language phrase by leveraging techniques used in semantic segmentation.
These methods condition a feed-forward segmentation network, such as a fully-convolutional network or U-Net, on the encoding of the natural language (\eg, LSTM over words).
The advantage is that these methods can be directly optimized for the
segmentation performance and can easily handle stuff categories as well as different numbers of target regions.
However, they are not as competitive on small-sized objects.
We compare a strong baseline of RMI~\cite{liu2017recurrent} on our dataset.
 

More state-of-the-art methods are based on a two-stage framework of region proposal and ranking. 
Significant innovations in techniques have been due to the improved techniques for object
detection (\eg, Mask R-CNN~\cite{he2017maskrcnn}) as well as language comprehension. 
Some earlier works~\cite{mao2016generation, yu2016modeling, nagaraja16refexp, hu2015natural,luo2017comprehension, rohrbach2016grounding, wang2015learning, liu2017referring, chen2017query, plummer2018conditional} adopt a joint image-language embedding model to rank object proposals according to their matching scores to the input expressions. More recent works improve the proposal generation~\cite{yu2018rethinking, chen2017query}, introduce attention mechanisms~\cite{deng2018accuatt, ye2019cross, akbari2019multi} for accurate grounding, or leverage week supervision from captions~\cite{xiao2017weakly, datta2019align2ground}.

The two-stage framework has also been further extended to modular
comprehension inspired by neural module networks~\cite{andreas2016neural}. For example,
Hu \etal~\cite{hu2017compositional} introduce a compositional modular network for better handling of attributes and relationships.
Yu \etal~\cite{yu2018mattnet} propose a modular attention network (MattNet) to factorize the referring task into separate ones for the noun phrase, location, and relationships. 
Liu \etal~\cite{liu2019improving} improves MattNet by removing easy and dominant words and regions to learn more challenging alignments.
Several recent works~\cite{zhang2018grounding, wang2019neighbourhood, YangSibei_2019_ICCV, Liu_2019_ICCV, Bajaj_2019_ICCV, dogan2019neural, bajaj2019g3raphground} also apply reasoning on graphs or trees for more complicated phrases.
These approaches have several appealing properties such as more detailed modeling of different aspects of language descriptions.
However, these techniques have been primarily evaluated on datasets with a closed set of categories, and often with ground-truth instances provided.

Sadhu \etal~\cite{Sadhu_2019_ICCV} proposes zero-shot grounding to handle phrases with unseen nouns. 
Our work emphasizes further on the large number of categories, attributes and relationships, providing supervision over these long-tailed concepts and more detailed and straightforward evaluation.

\vspace{-0.05in}
\section{The \vgp Dataset}
\begin{figure*}[h!]
\centering
\includegraphics[width=\linewidth]{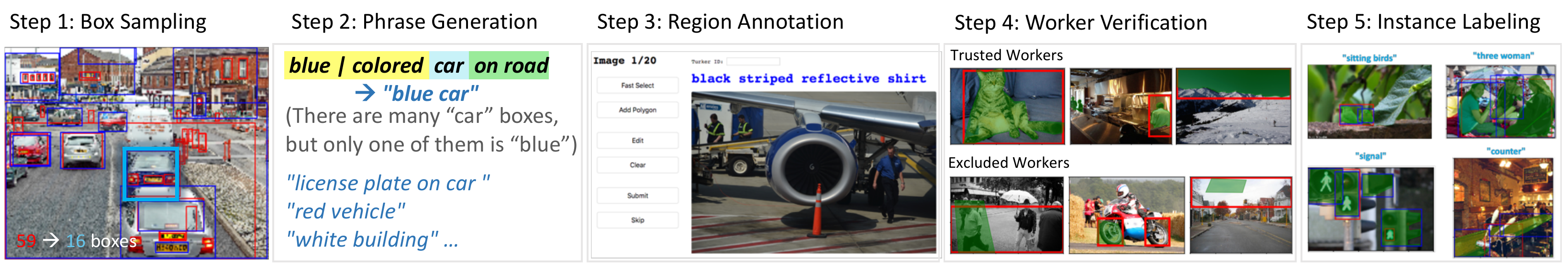}
\caption{\textbf{Illustrations of our \vgp dataset collection pipeline.} \textbf{Step 1:} blue boxes are the sampling result; red boxes are ignored.
\textbf{Step 2:} Phrase generation example in the previous image. \textbf{Step 3:} User interface for collecting region masks. \textbf{Step 4:} Example annotations from trusted and excluded annotators. 
\textbf{Step 5:} Instance label refinement examples. Blue boxes are final instance boxes, and red boxes are corresponding ones from Visual Genome annotations.}
\vspace{-0.1in}
\label{fig:collect}
\end{figure*}

\vspace{-0.05in}
\label{sec:data}
In this section, we describe how the \vgp dataset was collected, the statistics of the final annotations, and the evaluation metrics.
Our annotations are based on images and scene-graph annotations from the \textit{Visual Genome (VG)} dataset.  
We briefly describe each step in the data-collection pipeline illustrated in Figure~\ref{fig:collect}, deferring to the supplemental material Section~1.1 for more details.

%

\vspace{-0.18in}
\paragraph{Step 1: Box sampling} 
Each image in VG dataset contains 35 boxes on average, but they are highly redundant.
We sample an average of 5 boxes from each image in a stratified manner by avoiding boxes that are highly overlapping or are from a category that already has a high number of selected boxes. We also remove boxes that are less than 2\% or greater than 90\% of the image size.

\vspace{-0.18in}
\paragraph{Step 2: Phrase generation}
Each sampled box has \emph{several} annotations of category names (\eg, ``man'' and ``person''), attributes (\eg, ``tall'' and ``standing'') and relationships with other entities in the image (\eg, ``next to a tree'' and ``wearing a red shirt'').
We generate one phrase for one box at a time, by adding categories, attributes and relationships that allow discrimination
with respect to other VG boxes by the following set of heuristics:
\begin{compactenum}
\item We first examine if one of the provided categories of the selected box is unique.
If so we add this to the phrase and tack on to it a randomly sampled attribute or relationship description of the box.
The category name uniquely identifies the box in this image.
\item If the box is \emph{not} unique in terms of any of its category names, we look for a unique attribute of the box that distinguishes it from boxes of the same category.
If such an attribute exists we combine it with the category name as the generated phrase.
\item If \emph{no} such an attribute exists, we look for a distinguishing relationship description (a relationship predicate plus a category name for the supporting object). If such a relationship exists we combine it with the category name as the generated phrase.
\item If all of the above fail, we combine all attributes and relationships on the target box and randomly choose a category from the provided list of categories for the box to formulate the phrase.
In this case, the generated phrase is more likely to correspond to more than one instance within the image.
\end{compactenum}

The attribute and relationship information may be missing if the
original box does not have any, but there is always a category name for each box.
Phrases generated in this manner tend to be concise but do not always refer to a unique instance in the image.


\vspace{-0.18in}
\paragraph{Step 3: Region annotation} 
We present the images and generated phrases from the previous steps to human annotators on Amazon Mechanical Turk, and ask them to draw polygons around the regions that correspond to provided phrases. Around 10\% of phrases are skipped by workers when the phrases are ambiguous.

\vspace{-0.18in}
\paragraph{Step 4: Automatic annotator verification}
Based on manual inspection over a subset of annotators, we design an automatic mechanism to identify trusted annotators based on the overall agreement of their annotations with the VG boxes. 
Only annotations from trusted annotators are included in our dataset. 
9.27\% phrase-region pairs are removed in this step.



\vspace{-0.18in}
\paragraph{Step 5: Automatic instance labeling} As a final step we generate instance-level boxes and masks. 
In most cases, each polygon drawn by the annotators is considered an instance. It is further improved by a set of heuristics to merge multiple polygons into one instance and to split one polygon into several instances leveraging the phrase and VG boxes.



\begin{figure*}[t]
\centering
\includegraphics[width=\linewidth]{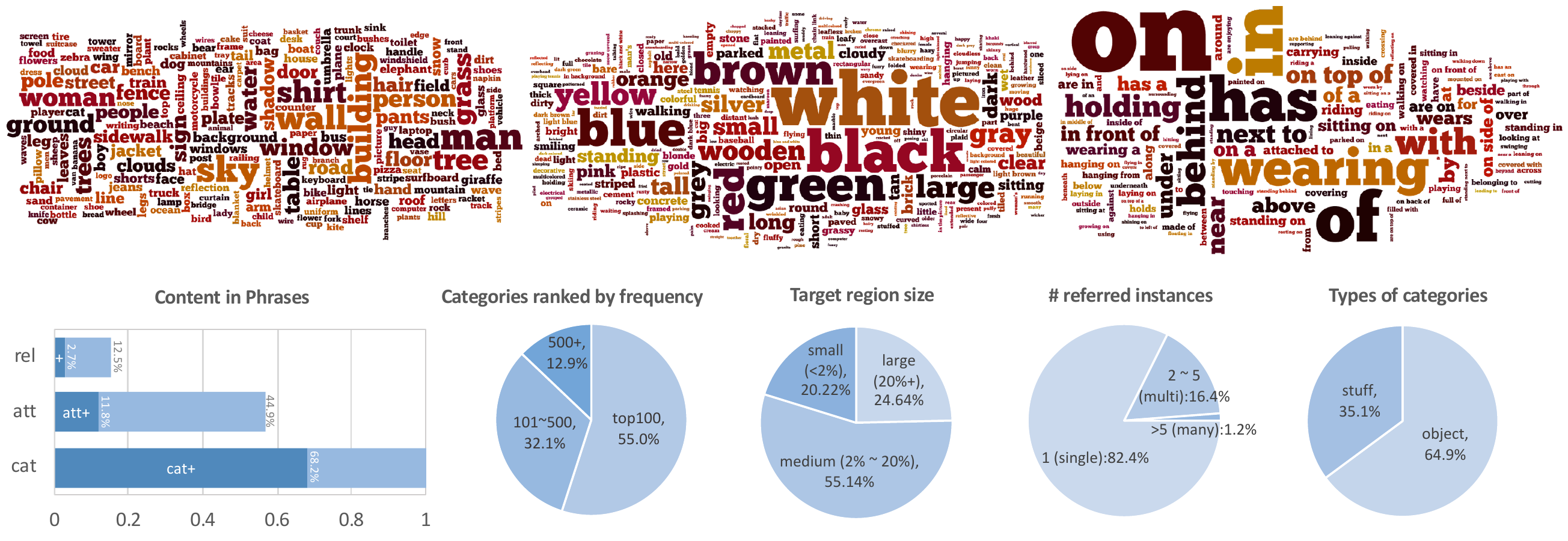}
\caption{\textbf{Statistics of the \vgp dataset}. 
  \textbf{Top row:} Word clouds of categories (left), attributes (center), and relationship descriptions (right) in the dataset. 
  The size of each phrase is proportional to the square root of its frequency in the dataset.
  \textbf{Bottom row:} breakdowns of the dataset into different subsets
  including contents in phrases (first), category frequency (second),
  size of target region relative to the image size (third),
  number of target instances per query phrase (fourth), and types of
  category (last). The leftmost bar chart shows the breakdown of
  phrases into those that have category annotation (cat) and
  those that can be distinguished by category information alone
  (cat+), and similarly for attributes and relationships. 
}
\vspace{-0.05in}
\label{fig:data_stat}
\end{figure*}

\subsection{Dataset statistics}
Our final dataset consists of 345,486 phrases across 77,262 images. 
This roughly covers 70\% of the images in Visual Genome.
We split the dataset into 310,816 phrases (71,746 images) 
for training, 20,316 (2,971 images) for validation, and 
14,354 (2,545 images) for testing. 
There is no overlap of COCO trainval images with our test split so that models pre-trained on COCO can be fairly used and evaluated.
Figure~\ref{fig:data_stat} illustrates several statistics of the dataset. 
Our dataset contains 1,272 unique category phrases, 593 unique attribute phrases, and 126 relationship phrases with frequency over 20, as seen by the word clouds.
Among the distribution of phrases (bottom left bar plot), one can see that 68.2\% of the instances can be distinguished by category alone (\emph{category+}), while 11.8\% of phrases require some treatment of attributes to distinguish instances (\emph{attributes+}).
Object sizes and their frequency vary widely.
While most annotations refer to a single instance, 17.6\% of phrases refer to two or more instances.
These aspects of the dataset make the \emph{PhraseCut} task challenging.
In Supplementary Section~1.2, we further demonstrate the long-tailed distribution of concepts and how attributes and relationships vary in different categories.

\subsection{Evaluation metrics}
\vspace{-0.03in}
The \emph{PhraseCut} task is to generate a
binary segmentation of the input image given a referring phrase. We assume that the input phrase is parsed into attribute, category, and relationship descriptions. 
For evaluation we use the following intersection-over-union (IoU) metrics:
\vspace{-0.05in}
\begin{itemize}
    \item cumulative IoU: $\texttt{cum-IoU}=\left(\sum_t I_t\right)/\left( \sum_t U_t\right)$, and
    \item mean IoU: $\texttt{mean-IoU}=\frac{1}{N}\sum_t{I_t}/{U_t}$. 
\end{itemize}
\vspace{-0.05in}
Here $t$ indexes over the phrase-region pairs in the evaluation set,  $I_t$ and  $U_t$ are the intersection and union area between predicted and ground-truth regions, and $N$ is the size of the evaluation set.
Notice that, unlike \texttt{cum-IoU}, \texttt{mean-IoU} averages the performance across all image-region pairs and thus balances the performance on small and large objects.

We also report the precision when each phrase-region task is considered correct if the \texttt{IoU} is above a threshold.
We report results with \texttt{IoU} thresholds at
0.5, 0.7, 0.9 as \texttt{Pr@0.5}, \texttt{Pr@0.7}, \texttt{Pr@0.9} respectively.

All these metrics can be computed on different subsets of the data to obtain a better understanding of the strengths and failure modes of the model.

\section{A Modular Approach to PhraseCut}
\vspace{-0.03in}
\label{sec:method}
\begin{figure}
\centering
\includegraphics[width=\linewidth]{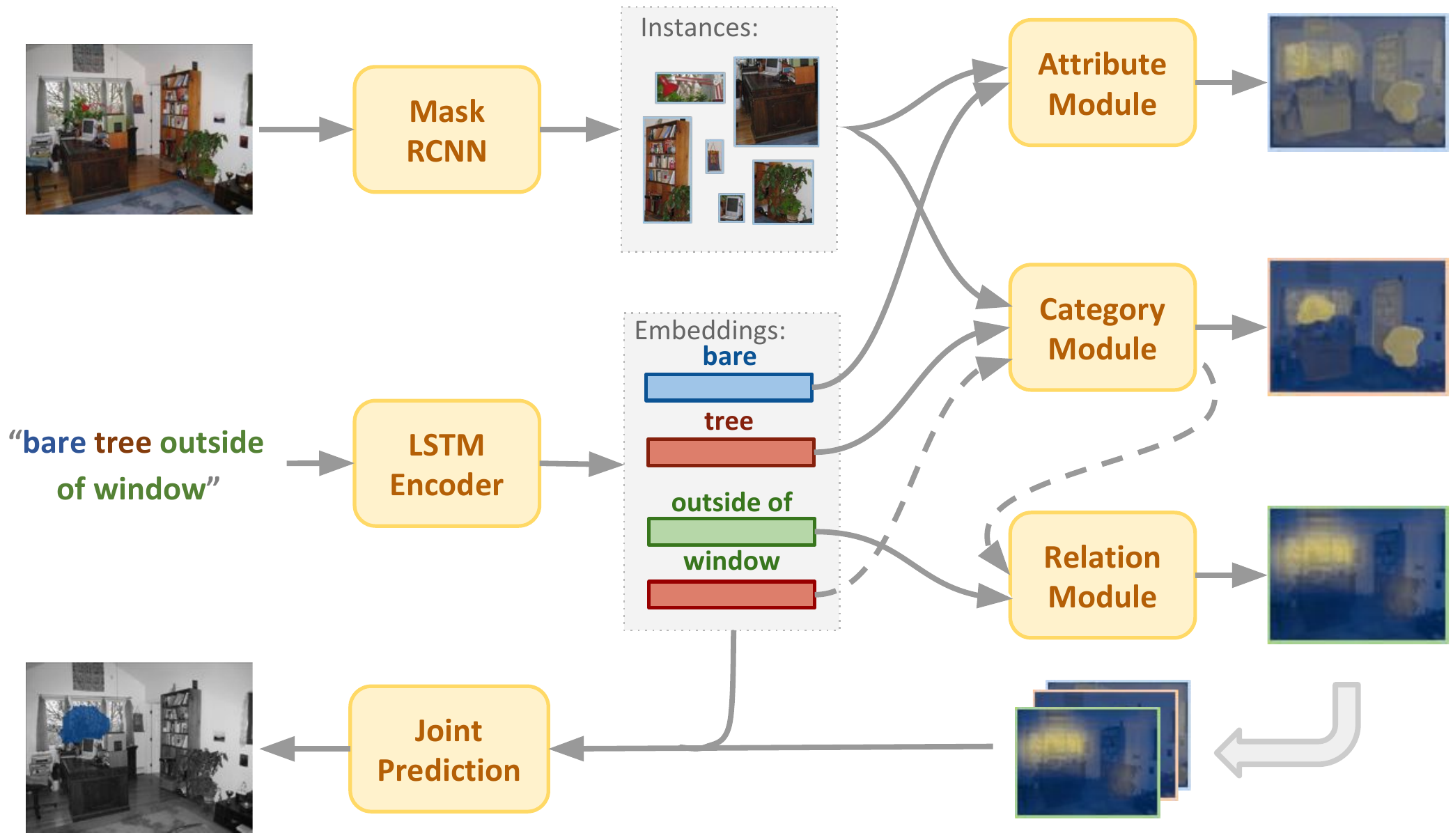}
\caption{\textbf{Architecture of HULANet}. The architecture consists of modules to obtain attribute, category, and relation predictions given a phrase and an image. The attribute and category scores are obtained from Mask-RCNN detections and projected back to the image. The scores across categories and attributes are combined using a module-specific attention model. The relationship module is a convolutional network that takes as input the prediction mask of the related category and outputs a spatial mask given the relationship predicate. The modules are activated based on their presence in the query phrase and combined using an attention mechanism guided by the phrase.}
\label{fig:model-e2e}
\end{figure}

We propose \textbf{H}ierarchical Mod\textbf{ul}ar \textbf{A}ttention \textbf{Net}work (HULANet) for the PhraseCut task, as illustrated in Figure~\ref{fig:model-e2e}.
The approach is based on two design principles.
First, we design individual modules for category, attribute and relationship sub-phrases.
Each module handles the long-tail distribution of concepts by learning to aggregate information across concepts using a module-specific attention mechanism.
Second, instance-specific predictions are projected onto the image space and combined using an attention mechanism driven by the input phrase.
This allows the model to handle stuff and object categories, as well as multiple instances in a unified manner.
Details of each module are described next.


\vspace{-0.18in}
\paragraph{Backbone encoders}
We use the Mask-RCNN~\cite{he2017maskrcnn} detector and bi-directional LSTMs~\cite{hochreiter1997long} as our backbone encoders for images and phrases respectively.
The Mask-RCNN (with ResNet101~\cite{he2016resnet} backbone) is trained to detect instances and predict category scores for the 1,272 categories that have a frequency over 20  on our dataset.
Different from instance detection tasks on standard benchmarks, we allow relatively noisy instance detections by setting a low threshold on objectness scores and by allowing at most 100 detections per image to obtain a high recall.
For phrase encoding, we train three separate bi-directional LSTMs to generate embeddings for categories, attributes and relationship phrases. 
They share the same word embeddings initialized from FastText~\cite{bojanowski2017enriching} as the input to the LSTM, and have mean pooling applied on the LSTM output of the corresponding words as the encoded output.

\vspace{-0.18in}
\paragraph{Category module}
The category module takes as input the phrase embedding of the category and detected instance boxes (with masks) from Mask-RCNN, and outputs a score-map of corresponding regions in the image.
We first construct the category channels $C \in \mathbb{R}^{N \times H \times W}$ by projecting the Mask-RCNN predictions back to the image.
Here $N=1272$ is the number of categories and $H \times W$ is set to $1/4\times$ the input image size.
Concretely, for each instance $i$ detected by Mask R-CNN as category $c_i$ with score $s_i$, we project its predicted segmentation mask to image as a binary mask $m_{i,H\times W}$, and update the category channel score at the corresponding location as $C[c_i, m_i] := \max(s_i, C[c_i, m_i])$. Finally, each category channel is passed though a ``layer-norm" which scales the mean and variance of each channel.

To compute the attention over the category channels, the phrase embedding ${e_{cat}}$ is passed through a few linear layers $f$ with sigmoid activation at the end to predict the attention weights over the category channels $A=\sigma(f(e_{cat}))$. We calculate the weighted sum of the category channels guided by the attention weights $S_{H\times W}=\sum_c A_c \cdot C_c$, and apply a learned affine transformation plus sigmoid to obtain the category module prediction heat-map $P_{H\times W}=\sigma(a\cdot S_{H\times W} + b)$.
This attention scheme enables the category module to leverage predictions from good category detectors to improve performance on more difficult categories. 
We present other baselines for combining category scores in the ablation studies in Section~\ref{sec:exp}.

\vspace{-0.18in}
\paragraph{Attribute module}
The attribute module is similar to the category module except for an extra attribute classifier. 
On top of the pooled ResNet instance features from Mask-RCNN, we train a two-layer multi-label attribute classifier. 
To account for significant label imbalance we weigh the positive instances more when training attribute classifiers with the binary cross-entropy loss.
To obtain attribute score channels we take the top 100 detections and project their top 20 predicted attributes back to the image. 
Identical with the category module, we use the instance masks from the Mask-RCNN, update the corresponding channels with the predicted attribute scores, and finally apply the attention scheme guided by the attribute embedding from the phrase to obtain the final attribute prediction score heat-map.


\vspace{-0.18in}
\paragraph{Relationship module}
Our simple relationship module uses the category module to predict the locations of the supporting object. 
The down-scaled ($32\times 32$) score of the supporting object is concatenated with the embedding of the relationship predicate. 
This is followed by two dilated convolutional layers with kernel size 7 applied on top, achieving a large receptive field without requiring many parameters.
Finally, we apply an affine transformation followed by sigmoid to obtain the relationship prediction scores.
The convolutional network can model coarse spatial relationships by learning filters corresponding to each spatial relation. For example, by dilating the mask one can model the relationship ``near", and by moving the mask above one can model the relationship ``on".

\vspace{-0.18in}
\paragraph{Combining the modules}
The category, attribute, and relation scores $P_{c}, P_{a}, P_{r}$ obtained from individual modules are each represented as a $H\times W$ image, $1/4$ the image size. 
To this we append channels of quadratic interactions $P_i \circ P_j$ for every pair of channels (including $i=j$), 
obtained using elementwise product and normalization, and a bias channel of all ones, to obtain a 10-channel scoremap~$F$ (3+6+1 channels).
Phrase embeddings of category, attribute and relationship are concatenated together and then encoded into 10-dimensional ``attention" weights $w$ through linear layers with LeakyReLU and DropOut followed by normalization.
When there is no attribute or relationship in the input phrase, the corresponding attention weights are set to zero and the attention weights are re-normalized to sum up to one.
The overall prediction is the attention-weighted sum of the linear and quadratic feature interactions: $O = \sum_t F_t w_t$.
Our experiments show a slight improvement of $0.05\%$ on validation \texttt{mean-IoU} with the quadratic features.

\vspace{-0.18in}
\paragraph{Training details}
The Mask-RCNN is initialized with weights pre-trained on the MS-COCO dataset~\cite{lin2014microsoft} and fine-tuned on our dataset. 
It is then fixed for all the experiments.
The attribute classifier is trained on ground-truth instances and their box features pooled from Mask-RCNN
with a binary cross-entropy loss specially weighted according to attribute frequency.
These are also fixed during the training of the referring modules.
On top of the fixed Mask-RCNN and the attribute classifier, we separately train the individual category and attribute modules.
When combining the modules we initialize the weights from individual ones and fine-tune the whole model end-to-end.
We apply a pixel-wise binary cross-entropy loss on the prediction score heat-map from each module and also on the final prediction heat-map. 
To account for the evaluation metric (\texttt{mean-IoU}), we increase the weights on the positive pixels and average the loss over referring phrase-image pairs instead of over pixels.
All our models are trained on the training set. 
For evaluation, we require a binary segmentation mask which is obtained by thresholding on prediction scores. 
These thresholds are set based on \texttt{mean-IoU} scores on the validation set.
In the next section, we report results on the test set.

\vspace{-0.03in}
\section{Results and Analysis}
\label{sec:exp}

\begin{table}
\begin{center}
\resizebox{0.9\linewidth}{!}{
\setlength{\tabcolsep}{2pt}
\begin{tabular}{l|c|c|c|c|c}
\textbf{Model}             & mean-IoU & cum-IoU & {Pr@0.5}  & Pr@0.7 & {Pr@0.9} \\
\hline
\textbf{HULANet} & & & & & \\
~~cat &39.9 &48.8	&40.8	&25.9	&5.5 \\
~~cat+att &\textbf{41.3} &\textbf{50.8} &\textbf{42.9} & \textbf{27.8} & \textbf{5.9} \\
~~cat+rel &41.1	&49.9	&42.3	&26.6	&5.6\\
	
~~cat+att+rel & \textbf{41.3} &50.2	&42.4	&27.0	&5.7 \\
\hline

\textbf{Mask-RCNN} self &36.2	&45.9	&37.2	&22.9	&4.1 \\

\textbf{Mask-RCNN} top &39.4	&47.4	&40.9	&25.8	&4.8\\
\hline

\textbf{RMI}  &21.1	&42.5	&22.0	&11.6	&1.5\\

\textbf{MattNet} &20.2	&22.7	&19.7	&13.5	&3.0
\end{tabular}
}
\end{center}
\vspace{-0.1in}
\caption{\textbf{Comparison of various approaches on the entire test set of \vgp.}
We compare different combinations of modules in our approach (HULANet) against baseline approaches: Mask-RCNN, RMI and MattNet.}
\vspace{-0.1in}
\label{tab:exp-overall}
\end{table}

\begin{table}
\begin{center}
\resizebox{0.9\linewidth}{!}{
\begin{tabular}{l|c|c|c|c|c}
\textbf{Model}            & all & coco & 1-100 & 101-500 & 500+ \\
\hline
\textbf{HULANet} & & & & & \\
~~cat &39.9	&46.5	&46.8	&31.8	&25.2\\
~~cat+att &\textbf{41.3}	&\textbf{48.3}	&\textbf{48.2}	&33.6	&26.6\\
~~cat+rel &41.1	&47.9	&47.8	&33.6	&26.6 \\
~~cat+att+rel &\textbf{41.3}	&47.8	&47.8	&\textbf{33.8}	&\textbf{27.1}\\
\hline
\textbf{Mask-RCNN} self &36.2	&44.9	&45.5	&27.9	&10.1\\
\textbf{Mask-RCNN} top &39.4 &46.1 &46.4 &31.6 &23.2\\
\hline
\textbf{RMI}  &21.1 &23.7 &28.4 &12.7 &5.5 \\
\textbf{MattNet} &20.2	&19.3	&24.9	&14.8	&10.6\\
\end{tabular}}
\end{center}
\vspace{-0.1in}
\caption{\textbf{The \texttt{mean-IoU} on \vgp test set for various category subsets.} The column \textit{coco} refers to the subset of data corresponding to the $80$ coco categories, while the remaining columns show the performance on the top 100, 101-500 and 500+ categories in the dataset sorted by frequency.}
\vspace{-0.1in}
\label{tab:exp-cat}
\end{table}

\begin{table}
\begin{center}
\resizebox{\linewidth}{!}{
\begin{tabular}{l|c|c|c|c|c|c|c}
\textbf{Model}            & all & att & att+ & rel & rel+ & stuff & obj \\
\hline
\textbf{HULANet} & & & & & & & \\
~~cat &39.9	&37.6	&37.4	&32.3	&33.0	&47.2	&33.9\\
~~cat+att &\textbf{41.3}	&\textbf{39.1}	&\textbf{38.8}	&33.7	&33.8	&\textbf{48.4}	&35.5\\
~~cat+rel &41.1	&38.8	&38.4	&33.8	&\textbf{34.0}	&48.1	&35.4\\
~~cat+att+rel &\textbf{41.3}	&39.0	&38.5	&\textbf{34.1}	&33.9	&48.3	&\textbf{35.6}\\
\hline
\textbf{Mask-RCNN} self &36.2	&34.5	&34.7	&29.0	&30.8	&44.4	&29.5\\
\textbf{Mask-RCNN} top  &39.4	&37.3	&36.6	&31.9	&32.6	&46.4	&33.6\\
\hline
\textbf{RMI} &21.1	&19.0	&21.0	&11.6	&12.2	&31.1	&13.0\\
\textbf{MattNet} &20.2	&19.0	&18.9	&15.6	&15.1	&25.5	&16.0\\
\hline
\hline
\textbf{Model}            &all & single & multi & many  & small & mid &large \\
\hline
\textbf{HULANet} & & & & & & & \\
~~cat & 39.9 & 41.2 & 37.0 & 34.3 & 15.1 & 40.3 & 67.6 \\
~~cat+att & \textbf{41.3 }& \textbf{42.6} & \textbf{38.6} & \textbf{35.9} & 17.1 & \textbf{42.0} & 68.0 \\
~~cat+rel & 41.1 & 42.5 & 38.2 & 35.5 & 17.1 & 41.5 & \textbf{68.2} \\
~~cat+att+rel & \textbf{41.3} & \textbf{42.6} & 38.4 & 35.7 & 17.3 & 41.7 & \textbf{68.2} \\
\hline
\textbf{Mask-RCNN} self & 36.2 & 37.2 & 34.1 & 29.9 & 17.0 & 35.7 & 59.4 \\
\textbf{Mask-RCNN} top & 39.4 & 40.6 & 36.8 & 33.4 & \textbf{18.5} & 39.3 & 63.6 \\
\hline
\textbf{RMI}  & 21.1 & 23.1 & 16.9 & 12.7 & 1.2  & 18.6 & 49.5 \\
\textbf{MattNet} & 20.2 & 22.2 & 15.9 & 12.6 & 6.1  & 18.9 & 39.5\\
\end{tabular}}
\end{center}
\vspace{-0.1in}
\caption{\textbf{The \texttt{mean-IoU} on \vgp test set for additional subsets.} \textit{att/rel}: the subset with attributes/relationship annotations; \textit{att+/rel+}: the subset which requires attributes or relationships to distinguish the target from other instances of the same category; \textit{single/multi/many}: subsets that contain different number of instances referred by a phrase; \textit{small/mid/large}: subsets with different sizes of the target region.}
\vspace{-0.1in}
\label{tab:exp-other}
\end{table}

\subsection{Comparison to baselines}
\vspace{-0.03in}
Table~\ref{tab:exp-overall} shows the overall performance of our model and its ablated versions with two baselines: RMI~\cite{liu2017recurrent} and MattNet~\cite{yu2018mattnet}. 
They yield near state-of-the-art performance on datasets such as RefCOCO~\cite{kazemzadeh2014referitgame}. 

RMI is a single-stage visual grounding method. It extracts spatial image features through a convolutional encoder, introduces convolutional multi-modal LSTM for jointly modeling of visual and language clues in the bottleneck, and predicts the segmentation through an upsampling decoder. 
We use the RMI model with ResNet101~\cite{he2016resnet} as the image encoder.
We initialized the ResNet with weights pre-trained on COCO~\cite{lin2014microsoft}, trained the whole RMI model on our training data of image region and referring phrase pairs following the default setting as in their public repository,
and finally evaluated it on our test set.

\begin{figure*}
\centering
\includegraphics[width=\linewidth]{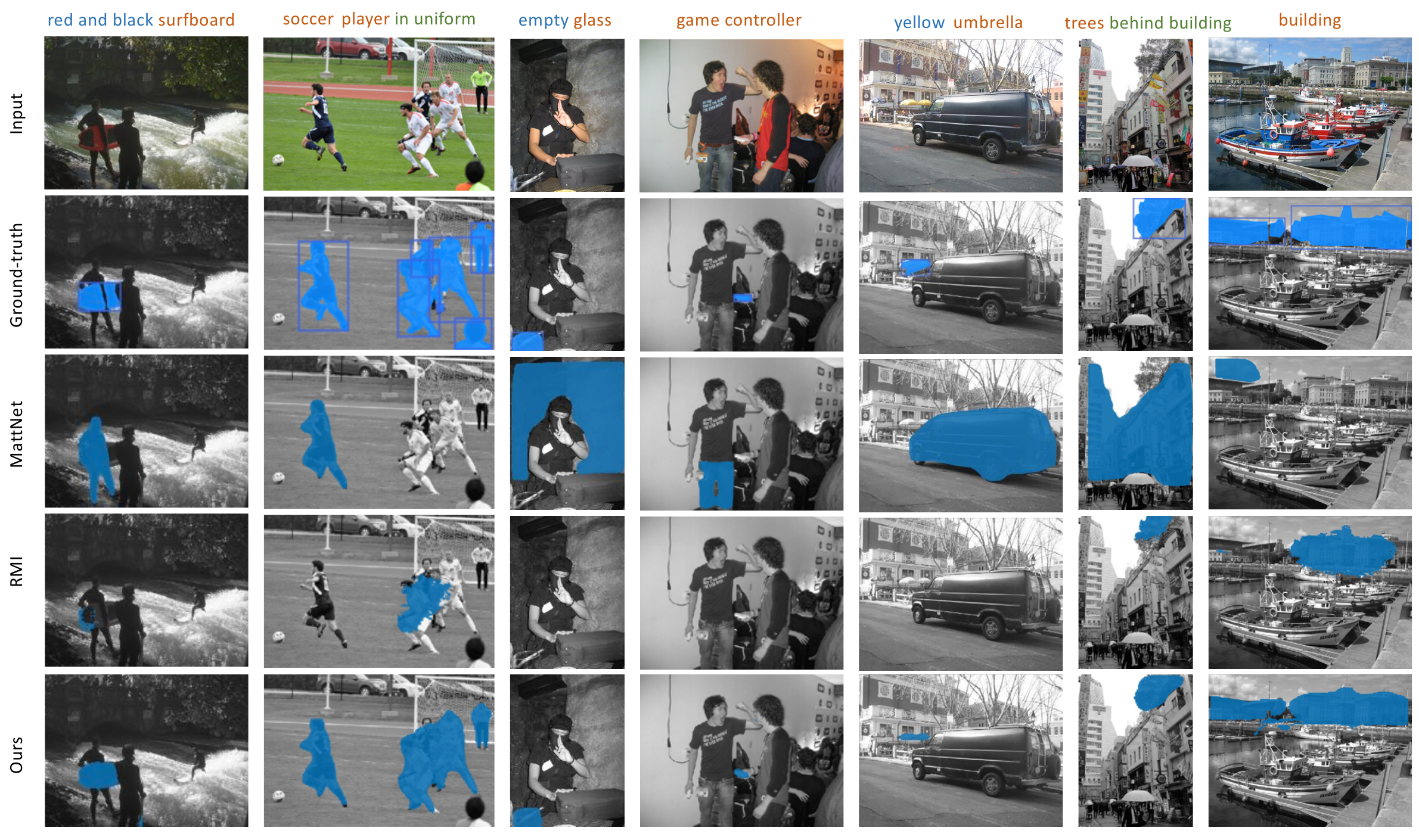}
\caption{\textbf{Prediction results on \vgp dataset.} Rows from top to down are:
(1) input image;
(2) ground-truth segmentation and instance boxes; 
(3) MattNet baseline; 
(4) RMI baseline; 
(5) HULANet (cat + att + rel).
See more results in the supplemental material. 
}
\label{fig:exp-example}
\end{figure*}

RMI obtains high \texttt{cum-IoU} but low \texttt{mean-IoU} scores because it handles large targets well but fails on small ones (see Table~\ref{tab:exp-other} ``small/mid/large" subsets). 
\texttt{cum-IoU} is dominated by large targets while our dataset many small targets: 
20.2\% of our data has the target region smaller than 2\% of the image area, while the smallest target in RefCOCO is 2.4\% of the image.
Figure~\ref{fig:exp-example} also shows that RMI predicts empty masks on challenging phrases and small targets.

MattNet focuses on ranking the referred box among candidate boxes.
Given a box and a phrase, it calculates the subject, location, and relationship matching scores with three separate modules, and predicts attention weights over the three modules based on the input phrase. 
Finally, the three scores are combined with weights to produce an overall matching score, and the box with the highest score is picked as the referred box. 

We follow the training and evaluation setup described in their paper.
We train the Mask-RCNN detector on our dataset, and also train MattNet to pick the target instance box among ground-truth instance boxes in the image.
Note that MattNet training relies on complete annotations of object instances in an image, which are used not only as the candidate boxes but also as the context for further reasoning. 
The objects in our dataset are only sparsely annotated, hence we leverage the Visual Genome boxes instead as context boxes.
At test time the top 50 Mask-RCNN detections from all categories are used as input to the MattNet model.

While this setup works well on RefCOCO, it is problematic on \vgp because detection is  more challenging in the presence of thousands of object categories.
MattNet is able to achieve \texttt{mean-IoU} = 42.4\% when the ground-truth instance boxes are provided in evaluation, but its performance drops to \texttt{mean-IoU} = 20.2\% when Mask-RCNN detections are provided instead.
If we only input the detections of the referred category to MattNet, \texttt{mean-IoU} improves to 34.7\%, approaching the performance of \textit{Mask-RCNN~self}, but it still performs poorly on rare categories. 

Our modular approach for computing robust category scores from noisy detections alone (\textit{HULANet cat}) outperforms both baselines by a significant margin.
Example results using various approaches are shown in Figure~\ref{fig:exp-example}.
Heatmaps from submodules and analysis of failure cases are included in Supplemental Section~3.

\vspace{-0.03in}
\subsection{Ablation studies and analysis}
\vspace{-0.05in}

Table~\ref{tab:exp-cat} shows that the performance is lower for rare categories.
Detection of thousands of categories is challenging, but required to support open-vocabulary natural language descriptions. 
However, natural language is also redundant.
In this section we explore if a category can leverage scores from related categories to improve performance, especially when it is rare.

First we evaluate Mask-RCNN as a detector, by using the mask of the top-1 detected instance from the referred category as the predicted region.
The result is shown as the row \textit{``Mask-RCNN self''} in Table~\ref{tab:exp-cat}. 
The row below  \textit{``Mask-RCNN top''} shows the performance of the model where each category is matched to a single other category based on the best \texttt{mean-IoU} on the training set. 
For example, a category ``pedestrian" may be matched to ``person" if the person detector is more reliable.
Supplemental Section~2 shows the full matching between source and target categories.
As one can see in Table~\ref{tab:exp-cat}, the performance on the tail categories jumps significantly (10.1\% $\rightarrow$ 23.2\% on the 500+ subset.)
In general the tail category detectors are poor and rarely used. 
This also points to a curious phenomenon in referring expression tasks where even though the named category is specific, one can get away with a coarse category detector. For example, if different animal species never appear together in an image, one can get away with a generic animal detector to resolve any animal species. 

This also explains the performance of the category module with the category-level attention mechanism. 
Compared to the single category picked by the Mask-RCNN top model, the ability of aggregating multiple category scores using the attention model provides further improvements for the tail categories. 
Although not included here, we find a similar phenomenon with attributes, where a small number of base attributes can support a larger, heavy-tailed distribution over the attribute phrases. 
It is reassuring that the number of visual concepts to be learned grows sub-linearly with the number of language concepts. 
However, the problem is far from solved as the performance on tail categories is still significantly lower.

Table~\ref{tab:exp-other} shows the results on additional subsets of the test data. 
Some high-level observations are that: (i) Object categories are more difficult than stuff categories. 
(ii) Small objects are extremely difficult. 
(iii) Attributes and relationships provide consistent improvements across different subsets.
Remarkably, the improvements from attributes and relationships are more significant on rare categories and small target regions where the category module is less accurate.

\vspace{-0.06in}
\section{Conclusion}
\vspace{-0.05in}
We presented a new dataset, \vgp, to study the problem of grounding natural language phrases to image regions.
By scaling the number of categories, attributes, and relations we found that existing approaches that rely on high-quality object detections show a dramatic reduction in performance.
Our proposed HULANet performs significantly better, suggesting that dealing with long-tail object categories via modeling their relationship to other categories, attributes, and spatial relations is a promising direction of research.
Another take away is that decoupling representation learning and modeling long-tails might allow us to scale object detectors to rare categories, without requiring significant amount of labelled visual data.
Nevertheless, the performance of the proposed approach is still significantly below human performance which should encourage better modeling of language and vision.

\vspace{-0.2in}
\paragraph{Acknowledgements} The project is supported in part by NSF Grants 1749833 and 1617917, and Faculty awards from Adobe. Our experiments were performed on the UMass GPU cluster obtained under the Collaborative Fund managed by the Massachusetts Technology Collaborative.


{\small
\bibliographystyle{ieee_fullname}
\bibliography{phrasecut}
}

\clearpage








\section*{Supplemental Material}
\setcounter{section}{0}
\setcounter{figure}{0}
\setcounter{table}{0}

The supplementary material provides details on the data collection pipeline and the long-tail distribution of concepts in the dataset. We also visualize more results from our proposed HULANet, including predictions from individual modules, failure cases, and comparison against baselines.

\section{\vgp Dataset}
\subsection{Further details on data collection}
As described in the main paper,
we start our data collection by mining Visual Genome (VG) boxes and phrases that are discriminative. We then
collect region annotations for each phrase from various human annotators. 
Human annotators are verified by comparing their annotations against VG boxes.
Finally, we merge and split collected regions to produce instance-level segmentation masks for each phrase.
The steps are described in more details below.

\vspace{-0.15in}
\paragraph{Step 1: Box sampling}
Each image in VG dataset contains an average of 35 boxes, many of which are redundant.
We sample a set of non-overlapping boxes across categories, also removing ones that are too small or large.

Define $r$ as the box size proportional to the image size. 
We ignore boxes with $r< 0.02$ or $r > 0.9$. 
For each image, we add the VG boxes to a sample pool one by one.
The current box is ignored if it overlaps a box already in the sample pool by \texttt{IoU}$>0.2$.
When sampling from the pool, the weight of each box being sampled is 
$w=\sqrt{\min(0.1, r)}$ so that we are less likely to get boxes with $r<0.1$.
Every time a new box is sampled, we divide the weights by $5$ for all boxes from the same category as the newly sampled box,
so that category diversity is encouraged.
Since the VG boxes are noisy, we only use annotations on these boxes to generate phrases, but not use the boxes as the ground-truth of the corresponding regions.

\begin{figure*}[h]
\centering
\includegraphics[width=0.85\linewidth]{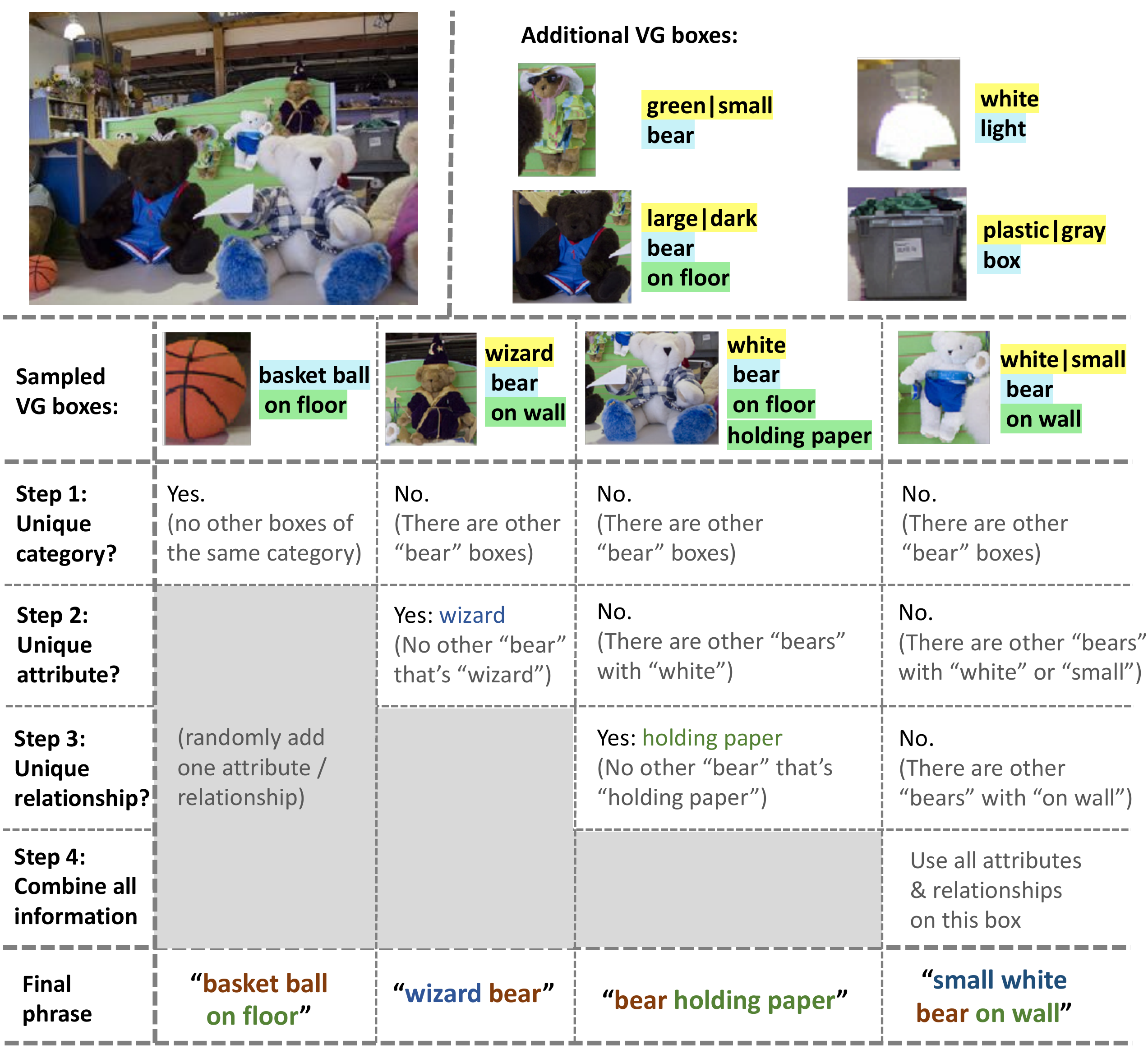}
\caption{\textbf{An illustrative example of phrase generation.} We show how we generate phrases from sampled Visual Genome~(VG) boxes in four steps. 
The raw image and additional VG boxes are displayed on top. We do not generate phrases on these additional VG boxes, but they are used to decide the uniqueness of sampled boxes.
VG annotations of categories, attributes and relationships on each box are highlighted in blue, yellow and green respectively.}
\label{fig:phrase}
\end{figure*}

\vspace{-0.15in}
\paragraph{Step 2: Phrase generation}
Figure~\ref{fig:phrase} shows an example of how we generate query phrases when collecting the \vgp dataset. The goal is to construct phrases that are concise, yet sufficiently discriminative of the target.

For VG boxes that have no other box from the same category in the image, the ``basket ball on floor" box for example, we randomly add an additional attribute / relationship annotation (if there is one) to the generated phrase. 
This avoids ambiguity caused by the missing VG box annotations, and makes it easier to find the corresponding regions, without making the phrase very long. 
Phrases generated this way are recognized as the ``cat+" subset in evaluation.

For VG boxes with unique attribute / relationship annotations within the same category, we generate the phrase by combining its  attribute / relationship annotation with the category name. 
In the ``wizard bear" and ``bear holding paper" examples, we obtain the phrase to refer to a single VG box, and avoided adding less helpful information (``on wall" or ``on floor") to the generated phrase.
They are recognized as the ``att+" and ``rel+" subsets in evaluation.

For the rest, we include all annotations we have on the sampled box into the generated phrase, like what we did in the ``small white bear on wall" example.
In these cases, the sampled VG boxes are usually more difficult to distinguish from other boxes, so we add all annotations to make the descriptions as precise as possible.
This is one of the sources of multi-region descriptions in our dataset. 

\vspace{-0.15in}
\paragraph{Step 3: Referred region annotation}
We present the images and phrases from the previous step to
human annotators on Amazon Mechanical Turk, and ask them to draw
polygons around the regions that correspond to those phrases.
In total we collected $383,798$ phrase-region pairs from $869$ different workers, which will be filtered in the next step.
In addition, we have $42,587$ phrases skipped by workers, $50.0\%$ with reason ``difficult to select", $24.8\%$ for ``wrong or ambiguous description", $23.7\%$ for ``described target not in image", and $1.5\%$ for ``other reasons".

\vspace{-0.15in}
\paragraph{Step 4: Automatic worker verification}
We designed an automatic worker verification metric in the spirit that statistically, annotations from better workers overlap more with corresponding VG boxes.

We rate each worker based on their overall agreement with VG boxes and the number of annotations they have done to identify a set of trusted workers.

We label a small set of annotations as ``good", ``so-so", or ``bad" ones, and  notice that the quality of annotations are strongly correlated with $\texttt{IoP}=s_\texttt{intersection} / s_\texttt{polygon}$ and $\texttt{IoU}=s_\texttt{intersection} / s_\texttt{union}$, 
where $s_\texttt{polygon}$ is the area of worker labeled polygons, $s_\texttt{intersection}$ and $s_\texttt{union}$ are the intersection and union between worker labeled polygons and VG boxes.
On the labeled set of annotations, we learn a linear combination of \texttt{IoP} and \texttt{IoU} that best separates ``good" and ``bad / so-so" annotations, defined as
$\texttt{agreement}= \texttt{IoP} + 0.8\times \texttt{IoU}$.

We calculate the average \texttt{agreement} score of all annotations
from each worker, and set a score threshold based on the total number of annotations from this worker:  
$\texttt{thresh}=\max(0.7, 0.95-0.05\times\#\text{annotations})$.
A worker is trusted if the average \texttt{agreement} score is above the \texttt{thresh}.
Workers with fewer than $10$ annotations are ignored.
Only annotations from trusted workers are included in our dataset.

In this step, $371$ out of $869$ workers are verified as trusted. $9.27\%$ (35,565 out of 383,798) phrase-region pairs are removed from our dataset.
In rare cases we have multiple annotations on the same input. We randomly select one of them, resulting in 345,486 phrase-region pairs.

\vspace{-0.15in}
\paragraph{Step 5: Automatic instance labeling from polygons}
Non-overlapping polygons are generally considered as separate instances with a few exceptions.
As a final step, we heuristically refine these instance annotations.

First, one instance can be split into multiple polygons when there are occlusions. If the category for a box is not plural and the bounding box of two polygons has a high overlap with it, then they are merged into a single instance.
Second, workers sometimes accidentally end a polygon and then create a new one to cover the remaining part of the same instance.
We merge two polygons into one instance if they overlap with each other, especially when one is much larger than the other.
Third, people tend to cover multiple instances next to each other with a single polygon. If a single polygon matches well with a set of VG boxes, these VG boxes are of similar sizes, and the referring phrase indicate plural instances, we split the polygon into multiple instances according to the VG boxes.

\begin{figure}[t]
	\centering
    \includegraphics[width=0.9\linewidth]{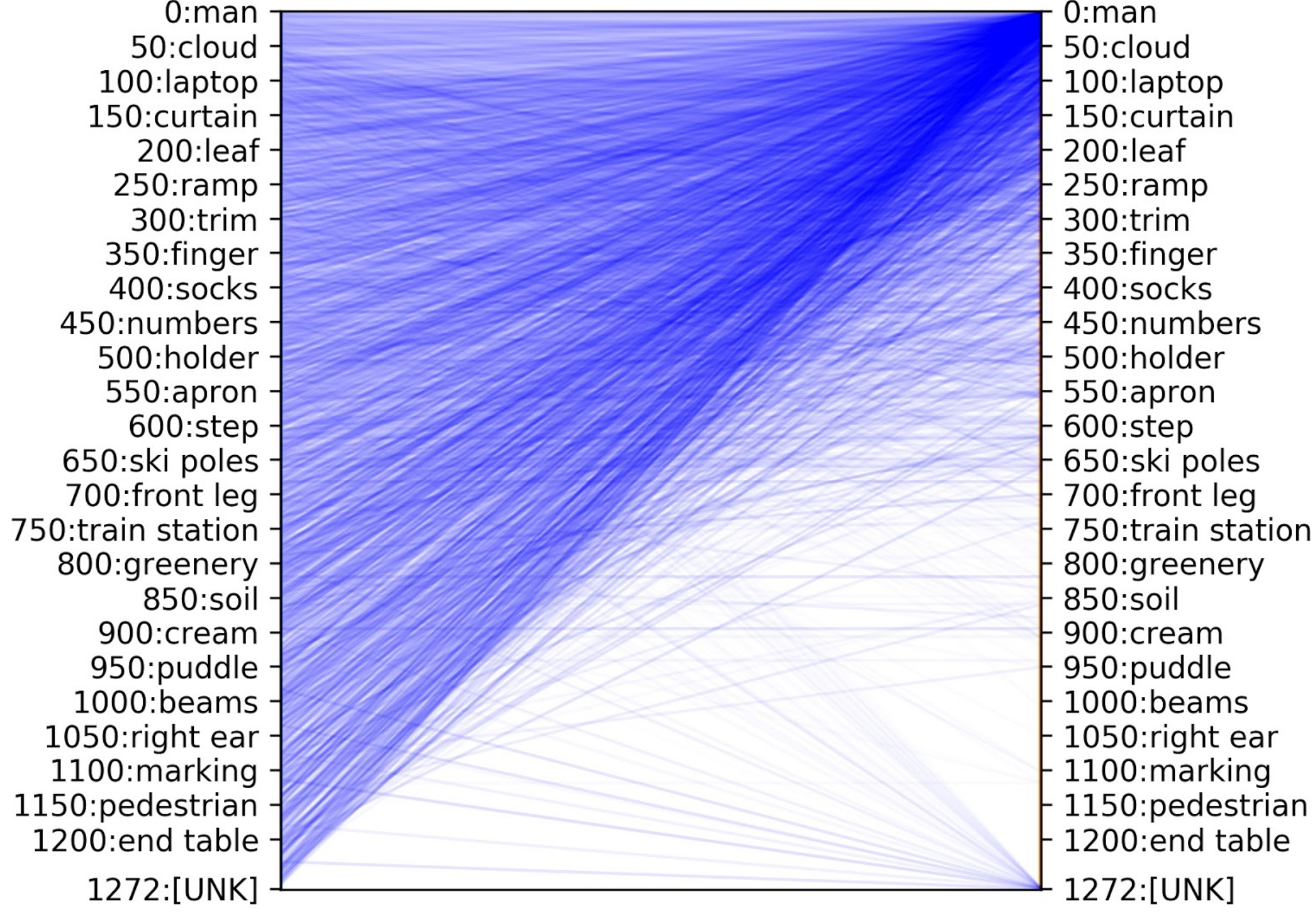}
	\caption{
		\textbf{Category matching in Mask-RCNN top.} Each input category (left)  are matched to its best substitute (right) measured by performance on training set. Categories are ordered from top to bottom by frequency.
	} 
	\label{fig:match}
\end{figure}

\begin{figure*}
\centering
\includegraphics[width=\linewidth]{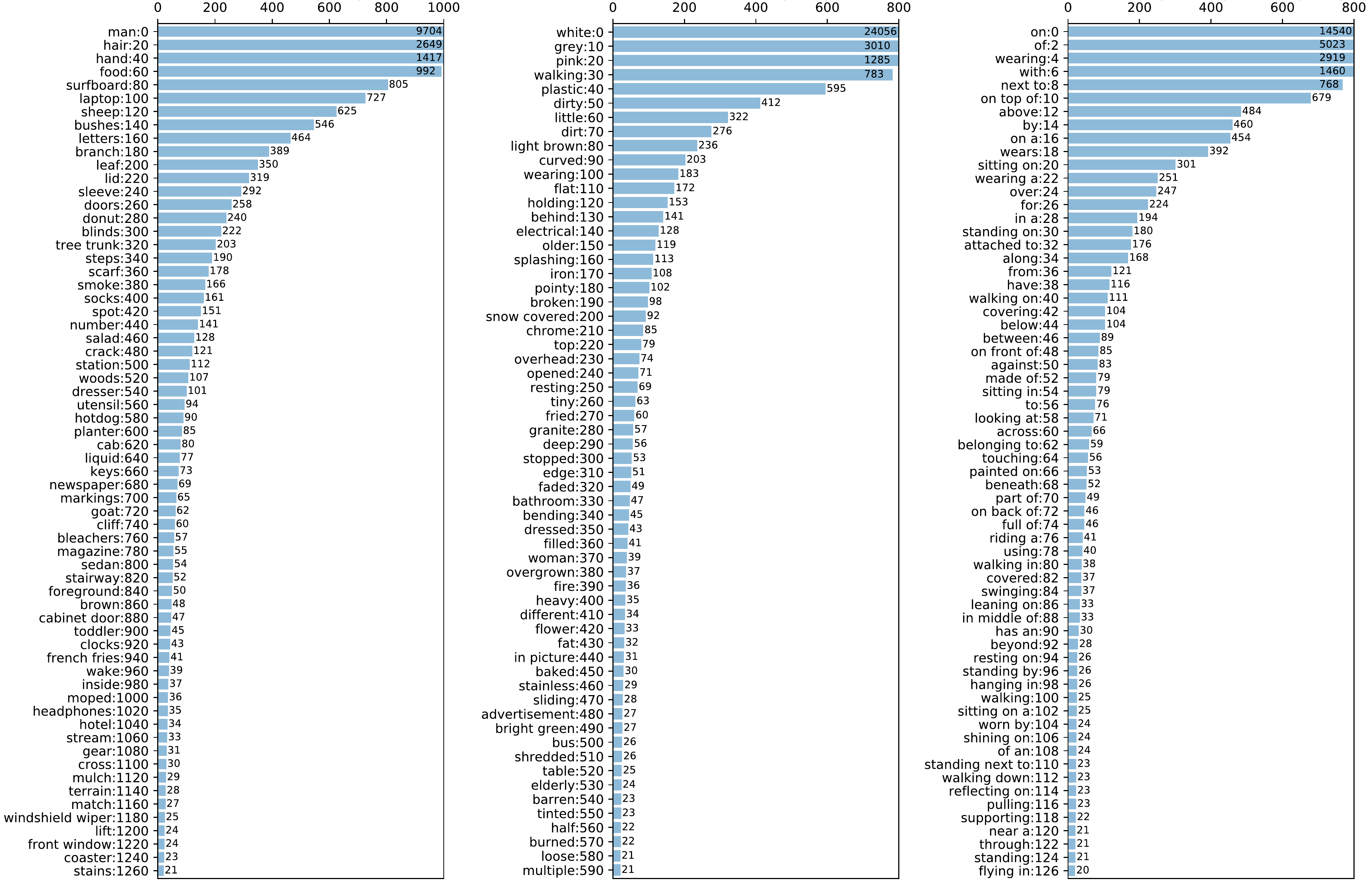}
  \caption{\textbf{Frequency histograms of categories (left), attributes (middle) and relationship descriptions (right).} Y-axis shows each entry and its frequency ranking; X-axis shows their frequency in the whole dataset.}
\label{fig:freq}
\end{figure*}

\begin{figure*}
\centering
\includegraphics[width=\linewidth]{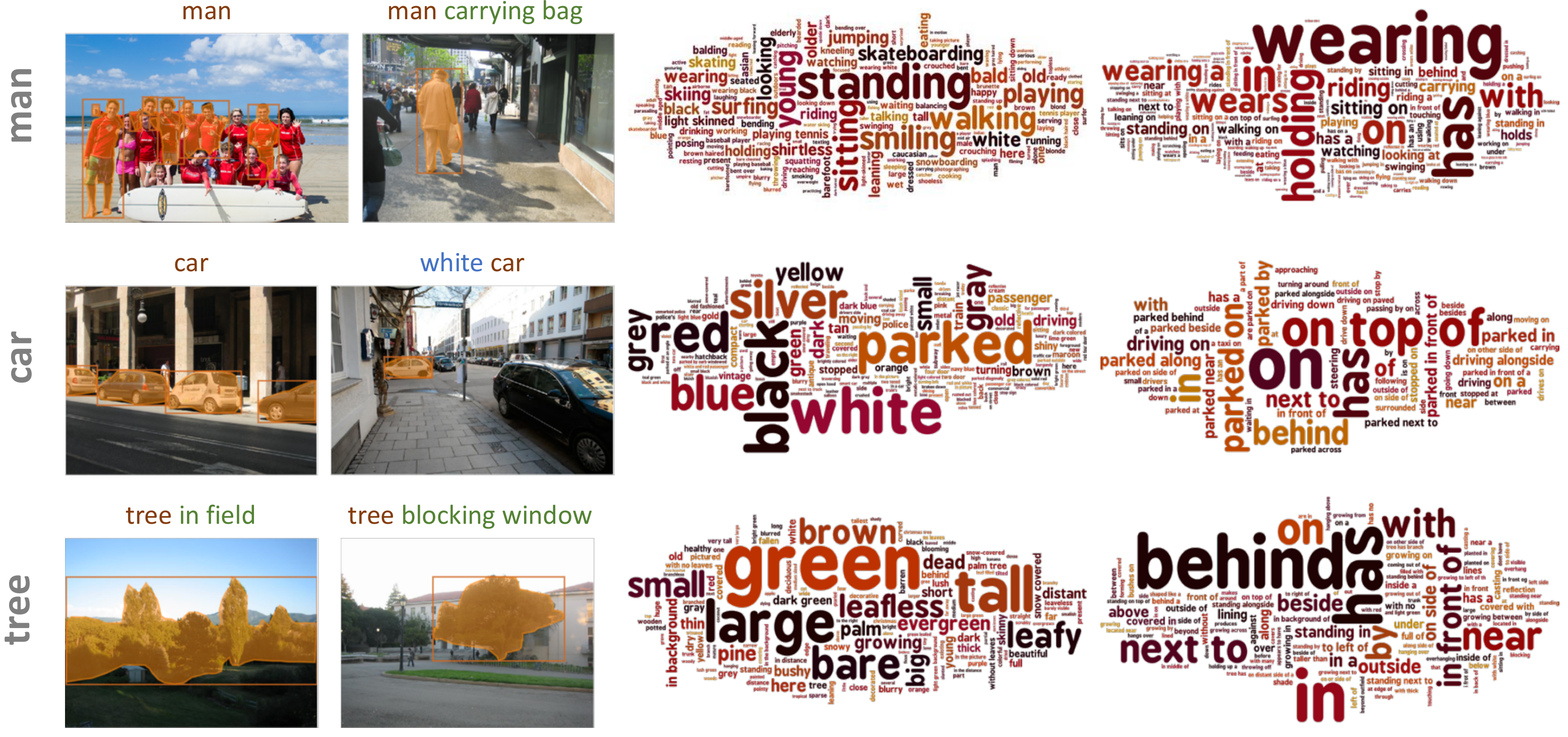}
  \caption{\textbf{Visualizations of ``man", ``car" and ``tree" categories.} From left two right, we display two examples, attribute cloud and relationship cloud within the given category. The size of each phrase is proportional to the square root of its frequency in the given category.}
\label{fig:cat}
\end{figure*}

\subsection{Additional dataset visualizations}
\vspace{-0.05in}

Figure~\ref{fig:freq} shows the frequency histograms for categories, attributes and relationship descriptions. 
Compared with tag clouds, the histograms better reveal the long-tail distribution of our dataset.

Figure~\ref{fig:cat} provides detailed visualizations for three typical categories: ``man", ``car" and ``tree".
The attribute and relationship description distributions vary a lot for different categories. 
Attributes and relationships mainly describe clothing, states and actions for ``man".
In the ``car" category, attributes are focused on colors, while relationships are about locations and whether the car is parked or driving.

We can see several sets of opposite concepts in ``tree" attributes, such as ``large - small", ``green - brown", ``bare/leafless - leafy", ``growing - dead", etc.
Relationships for ``tree" mainly describe the locations.

\section{Category Matching in Mask-RCNN top}
In \textit{Mask-RCNN top} we map each input category to its substitute category.
Given an input category, we consider every referring phrase in the training set containing this category, and pick the best category with which the detected mask yields highest \texttt{mean-IoU} on each referring phrase.
The final substitute category is the one that most frequently picked as the best.
The mappings are shown in Figure~\ref{fig:match}.
Using detections of a related and frequent category is often better.
Detections from categories with frequency ranked beyond 600 are rarely used.

\section{Additional Results from HULANet}

\subsection{Modular heatmaps}
\vspace{-0.05in}

\begin{figure*}
\centering
\includegraphics[width=0.75\linewidth]{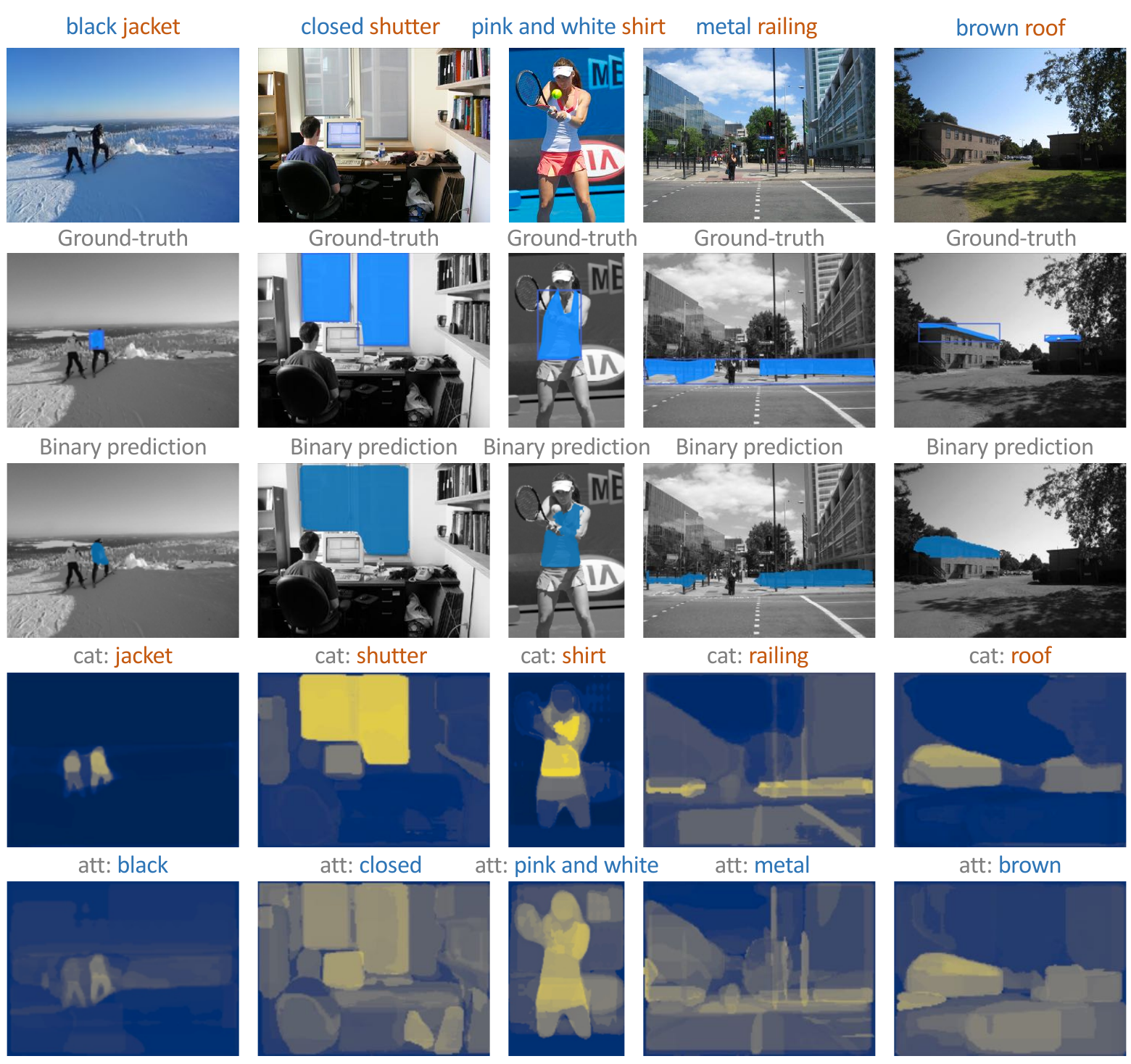}
\caption{\textbf{HULANet prediction results and heatmaps on phrases with attributes.} 
Rows from top to down are: 
(1) input image;
(2) ground-truth segmentation and instance boxes; 
(3) predicted binary mask from HULANet (cat+att+rel);
(4) heatmap prediction from the category module;
(5) heatmap prediction from the attribute module.}
\label{fig:att}
\end{figure*}

\begin{figure*}
\centering
\includegraphics[width=0.65\linewidth]{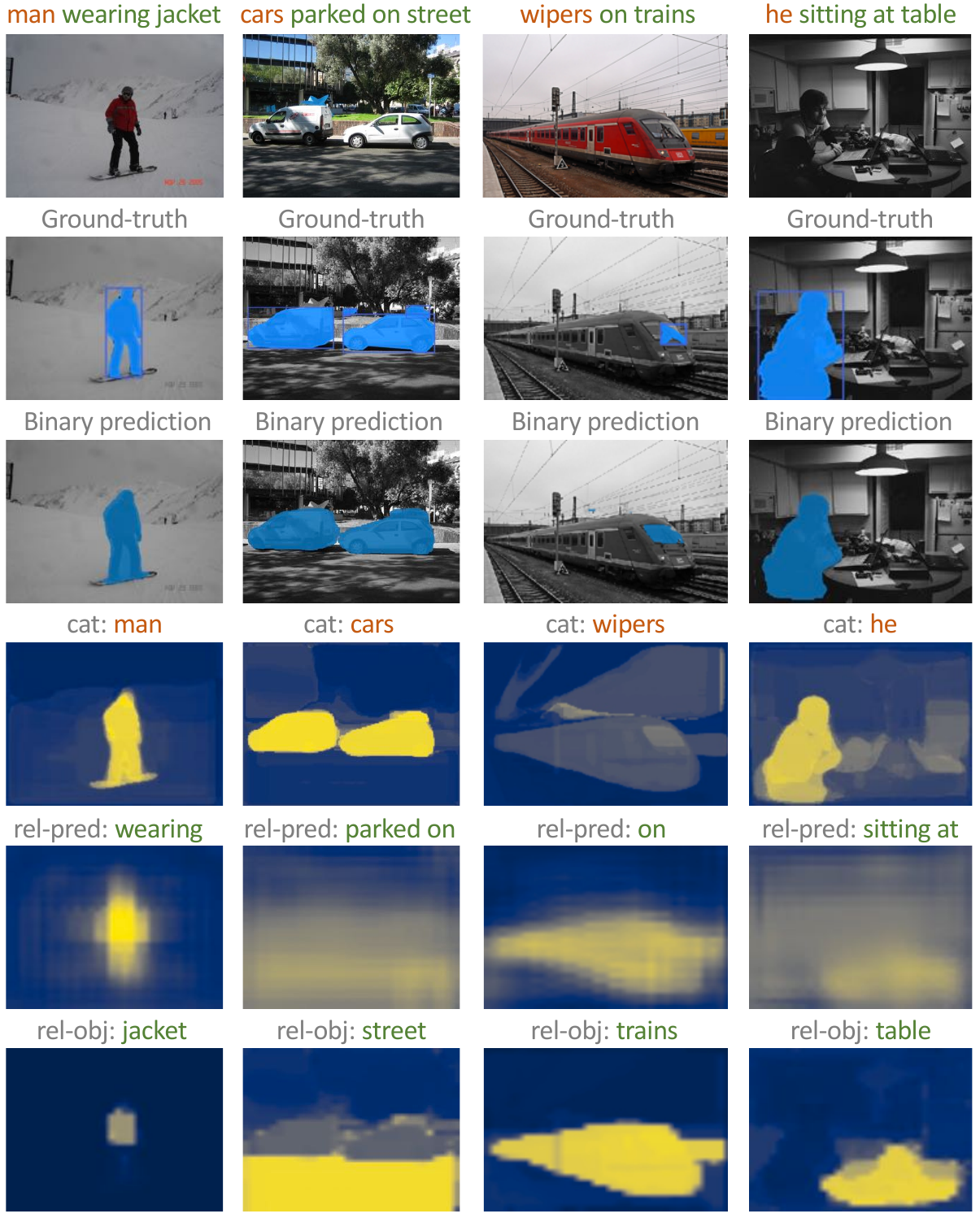}
\caption{\textbf{HULANet prediction results and heatmaps on phrases with relationships.} 
Rows from top to down are: 
(1) input image;
(2) ground-truth segmentation and instance boxes; 
(3) predicted binary mask from HULANet (cat+att+rel);
(4) heatmap prediction from the category module;
(5) heatmap prediction from the relationship module;
(6) heatmap prediction of the supporting object (in the relationship description) from the category module.}
\label{fig:rel}
\end{figure*}

In Figure~\ref{fig:att} and Figure~\ref{fig:rel}, we show HULANet predictions and modular heatmaps. 

Figure~\ref{fig:att} demonstrates that our attribute module is able to capture color (``black", ``brown"), state (``closed"), material (``metal") and long and rare attributes (``pink and white"). In the first (``black jacket") example, the category module detects two jackets, while the attribute module is able to select out the ``black" one against the white one.

Figure~\ref{fig:rel} shows how our relationship module modifies the heatmaps of supporting objects depending on different relationship predicates. 
With the predicate ``wearing", the relationship module predicts expanded regions of the detected ``jacket" especially vertically. 
The relational prediction of ``parked on" includes regions of the ``street" itself as well as regions directly above the ``street", while the predicate ``on" leads to the identical region prediction as the supporting object.
In the last example of ``sitting at", a broader region around the detected ``table" is predicted, covering almost the whole image area.

\subsection{Failure case analysis}
\begin{figure*}
\centering
\includegraphics[width=0.95\linewidth]{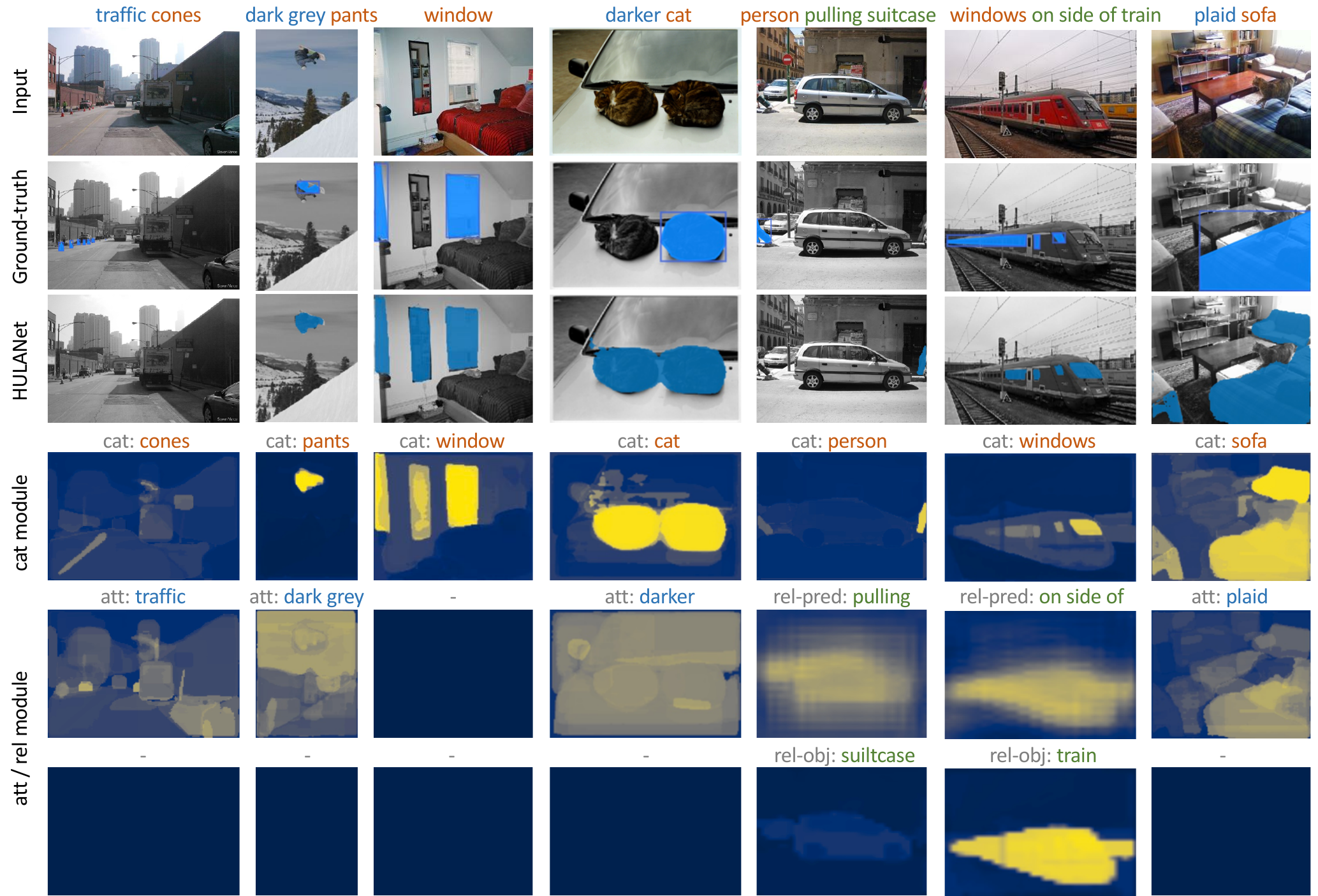}
\caption{\textbf{Negative results from HULANet on \vgp test set.} Rows from top to down are: 
(1) input image;
(2) ground-truth segmentation and instance boxes; 
(3) predicted binary mask from HULANet (cat+att+rel);
(4) heatmap prediction from the category module;
(5-6) heatmap predictions from additional (attribute or relationship) modules.}
\vspace{-0.1in}
\label{fig:neg}
\end{figure*}

\vspace{-0.05in}
Figure~\ref{fig:neg} displays typical failure cases from our proposed HULANet. Heatmaps from internal modules provide more insights where and why the model fails.

In the first example, our backbone Mask-RCNN fails to detect the ground-truth ``traffic cones", which are extremely small and from rare categories.
Similarly, in the second ``dark grey pants" example, the ``pants" is not detected as a separate instance in the backbone Mask-RCNN, therefore the category module can only predict the whole mask of the skateboarder.

The third ``window" example shows when the category module (and the backbone Mask-RCNN) fails to distinguish mirrors from windows.
In the fourth example, our attribute module fails to recognize which cat is ``darker" than the other.

We then display two failure cases for the relationship module. It fails on the first one because the supporting object (``suitcase") is not detected by the category module, and fails on the second one for unable to accurately model the relation predicate ``on side of".

In the last example, although our attribute module figures out which sofa is ``plaid", the final prediction is dominated by the category module and fails to exclude non-plaid sofas.

\subsection{More comparisons against baseline methods}
\begin{figure*}
\centering
\includegraphics[width=0.95\linewidth]{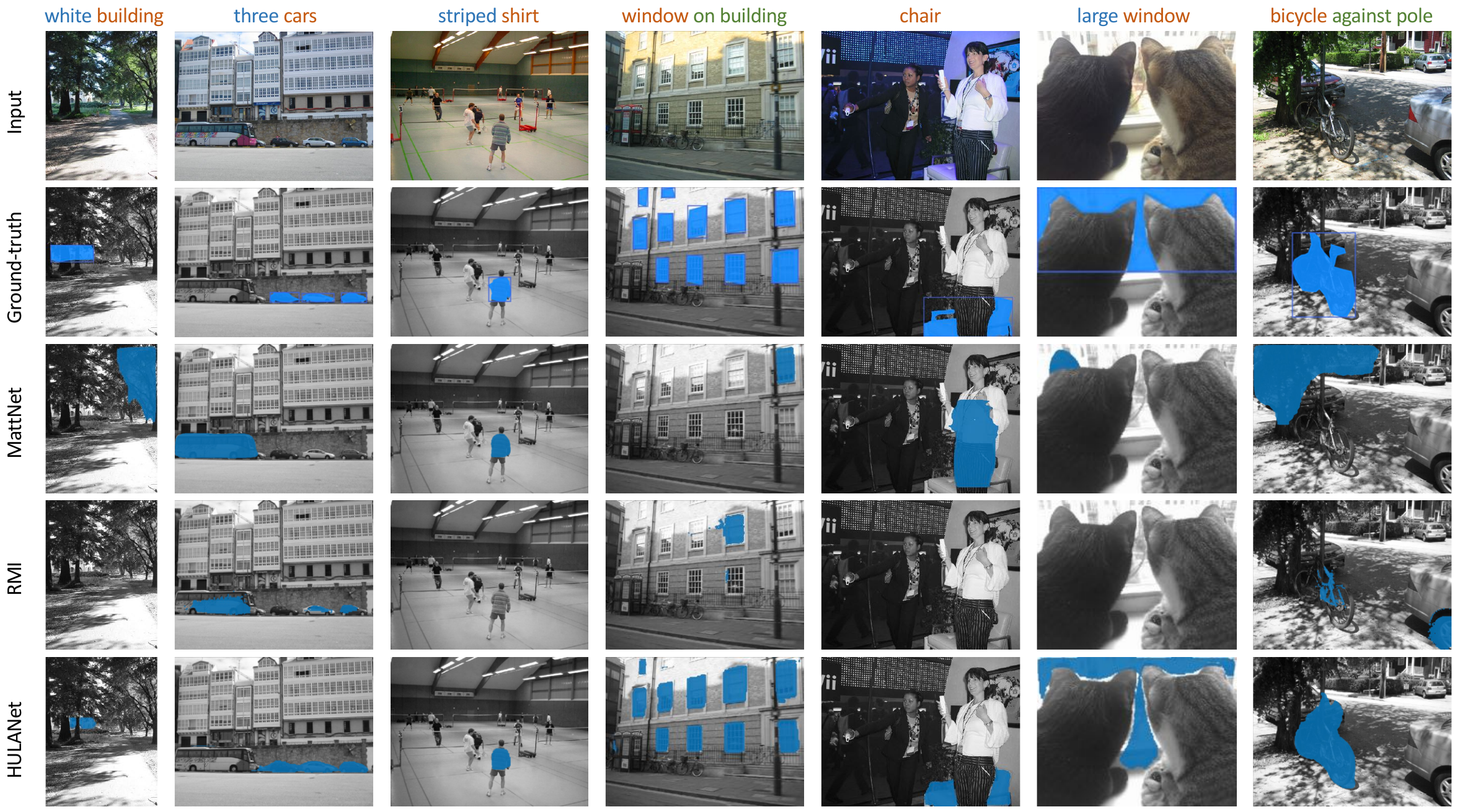}
\caption{\textbf{More prediction results comparing against baseline models on \vgp test set.} Rows from top to down are: 
(1) input image;
(2) ground-truth segmentation and instance boxes; 
(3) MattNet baseline; 
(4) RMI baseline; 
(5) HULANet (cat + att + rel).}
\vspace{-0.1in}
\label{fig:result}
\end{figure*}

\vspace{-0.05in}
As an extension to Figure~7 in the main paper, Figure~\ref{fig:result} here shows more examples of prediction results compared against baseline methods.
The ``white building", ``chair" and ``large window" examples demonstrate that our HULANet is better at handling occlusions.

\end{document}